\documentclass[preprint, 3p,
authoryear]{elsarticle} 

\usepackage[hyphens]{url}

\usepackage{graphicx}

\usepackage[T1]{fontenc}
\usepackage{lmodern}
\usepackage{amssymb,amsmath}
\usepackage{lineno} 
\usepackage{ifxetex,ifluatex}
\usepackage{fixltx2e} 
\IfFileExists{upquote.sty}{\usepackage{upquote}}{}
\ifnum 0\ifxetex 1\fi\ifluatex 1\fi=0 
  \usepackage[utf8]{inputenc}
\else 
  \usepackage{fontspec}
  \ifxetex
    \usepackage{xltxtra,xunicode}
  \fi
  \defaultfontfeatures{Mapping=tex-text,Scale=MatchLowercase}
  
\fi
\IfFileExists{microtype.sty}{\usepackage{microtype}}{}
\usepackage[]{natbib}
\ifxetex
  \usepackage[setpagesize=false, 
              unicode=false, 
              xetex]{hyperref}
\else
  \usepackage[unicode=true]{hyperref}
\fi
\hypersetup{breaklinks=true,
            bookmarks=true,
            pdfauthor={},
            pdftitle={An empirical evaluation of imbalanced data strategies from a practitioner's point of view},
            colorlinks=false,
            urlcolor=blue,
            linkcolor=magenta,
            pdfborder={0 0 0}}

\providecommand{\tightlist}{%
  \setlength{\itemsep}{0pt}\setlength{\parskip}{0pt}}

\usepackage[utf8]{inputenc}
\usepackage[T1]{fontenc}
\usepackage{amsmath}
\usepackage{subfig}
\usepackage{floatrow}
\usepackage{bookmark}
\usepackage{graphicx}
\usepackage[english]{babel}
\newfloatcommand{capbtabbox}{table}[][\FBwidth]
\def\tightlist{}
\usepackage{booktabs}
\usepackage{longtable}
\usepackage{array}
\usepackage{multirow}
\usepackage{wrapfig}
\usepackage{float}
\usepackage{colortbl}
\usepackage{pdflscape}
\usepackage{tabu}
\usepackage{threeparttable}
\usepackage{threeparttablex}
\usepackage[normalem]{ulem}
\usepackage{makecell}
\usepackage{xcolor}

\begin{document}

\begin{frontmatter}

  \title{An empirical evaluation of imbalanced data strategies from a
practitioner's point of view}
    \author[a]{Jacques Wainer%
  \corref{cor1}%
  }
   \ead{wainer@ic.unicamp.br} 
      \affiliation[a]{Artificial Intelligence Lab, REOD.ai, Institute of
Computing, University of Campinas, Brazil}
    \cortext[cor1]{Corresponding author}
  
  \begin{abstract}
  This paper evaluates six strategies for mitigating imbalanced data:
  oversampling, undersampling, ensemble methods, specialized algorithms,
  class weight adjustments, and a no-mitigation approach referred to as
  the baseline. These strategies were tested on 58 real-life binary
  imbalanced datasets with imbalance rates ranging from 3 to 120. We
  conducted a comparative analysis of 10 under-sampling algorithms, 5
  over-sampling algorithms, 2 ensemble methods, and 3 specialized
  algorithms across eight different performance metrics: accuracy, area
  under the ROC curve (AUC), balanced accuracy, F1-measure, G-mean,
  Matthew's correlation coefficient, precision, and recall.
  Additionally, we assessed the six strategies on altered datasets,
  derived from real-life data, with both low (3) and high (100 or 300)
  imbalance ratios (IR).

  The principal finding indicates that the effectiveness of each
  strategy significantly varies depending on the metric used. The paper
  also examines a selection of newer algorithms within the categories of
  specialized algorithms, oversampling, and ensemble methods. The
  findings suggest that the current hierarchy of best-performing
  strategies for each metric is unlikely to change with the introduction
  of newer algorithms.
  \end{abstract}
  
 \end{frontmatter}

\hypertarget{introduction}{%
\section{Introduction}\label{introduction}}

In a class imbalance binary classification problem, the proportion of
examples from each class is not equal in the training data and
potentially in the test or future data. The minority class is typically
referred to as the \textbf{positive} class, while the majority class is
the \textbf{negative} class. The imbalance rate (IR) represents the
proportion of the majority class relative to the minority class. It is
worth noting that sometimes the inverse of this ratio is used as the
imbalance ratio. Fortunately, these two distinct uses can be easily
identified. Although there is no universally accepted threshold for when
the data set is \emph{imbalanced}, the published research on imbalanced
data sets typically focuses on imbalance rates ranging from 3 to 100.

Imbalanced classification problems pose two main challenges. Firstly,
detecting the positive class is often more crucial than detecting the
negative class. In scenarios like disease detection, misclassifying a
person as healthy when they have the disease is more serious than
misclassifying a healthy person as having the disease. This is because
positive cases require further analysis, unlike false negatives who will
not seek treatment since they have been diagnosed as healthy.

Secondly, the scarcity of positive class examples can lead to suboptimal
classification results. For instance, when minimizing a global metric,
it may be reasonable to prioritize minimizing errors for the majority
class while accepting errors for the positive class.

The problem of bias for the negative class can be worsened in local
regions, as discussed by \citet{prati2004class},
\citet{lopez2013insight}, and \citet{haixiang2017learning}.
\emph{Intrinsic characteristics} of the data distribution play a
significant role in suboptimal classifiers. These characteristics can be
understood by distinguishing \emph{safe}, \emph{borderline}, and
\emph{noisy} data \citep{kubat1997addressing}. Safe data points reside
in homogeneous regions of their respective class, while noisy data
points are a small set of data points located in regions safe for the
other class. Borderline data points are close to the class boundary
where some overlap is expected. Classifiers may perform well in defining
safe regions, disregard noisy data, but differ in their treatment of
borderline regions. Imbalanced problems may result in a conceptual shift
from safe regions to borderline regions to noisy regions due to the
scarcity or low density of minority class data. This can lead to
different classifiers treating them differently, such as treating a
borderline region as a safe region for the majority class or considering
minority examples in this region as noisy data. Some of the intrinsic
characteristics discussed by \citet{lopez2013insight} can be understood
within this framework of safe, borderline, and noisy data. \emph{Small
disjuncts} \citep{jo2004class} refer to isolated clusters of
positive/minority class examples that may be challenging for classifiers
to learn. \emph{Low density} \citep{lopez2013insight} refers to positive
examples spread over large volumes, resulting in lower densities and
potentially being labeled as noise. \emph{Overlapping}
\citep{denil2010overlap, prati2004class} regions have equal numbers of
positive and negative examples, leaving the classifier with insufficient
information to assign different probabilities to the classes.
\emph{Noisy data} and \emph{borderline examples}
\citep{napierala2010learning} represent instances where safe positive
data points may be misinterpreted as noise or exist on the class
boundary, posing a higher risk of overlap.

The issue of \emph{data set shift} represents a more nuanced problem,
not directly linked to the safe, borderline, and noisy regions. A
fundamental assumption in machine learning is that the test set, on
which measurements are performed to make decisions about algorithmic
choices, is a valid approximation of future data. However, if there are
only a few positive examples, it is improbable that different test sets
and future data will be ``similar'' to each other, particularly
concerning the properties of the minority class. Consequently, decisions
made by the practitioner based on one test set may not be the most
effective when considering the data the system will encounter in the
future.

There has been an extensive literature proposing different
strategies/techniques/algorithms to deal with imbalanced data. In this
paper we will use the term \textbf{strategy} to refer to them all. Most
of the strategies discusses in this paper are full fledged algorithms
specially the sampling and ensemble strategies, and we will use the term
\textbf{algorithm} to discuss particular techniques within these
families. But some of the strategies are as simple as doing noting to
the data or the classifier (the strategy we call \textbf{baseline} in
this paper), while other are to change the base classifier algorithms so
that they weight false positive errors more than false negative (a
strategy we call \textbf{class weight}). It is a stretch to call these
alternatives as algorithms, and thus our use of the term strategy.

For the purposes of his paper, we will classify the different strategies
into the following groups or \textbf{families}:

\begin{itemize}
\item
  \textbf{Baseline} This strategy is to apply a ``standard'', unmodified
  classifier to the data. The standard classifier is called a
  \textbf{base classifier}.
\item
  \textbf{Cost sensitive:} These strategies add a cost to the
  classification construction algorithm or optimization so that errors
  for the majority and minority classes have different costs, and thus,
  the classifier would ``try harder'' to predict the minority class. A
  simple solution is to weight the data differently by classes, but the
  different classifier algorithms must be adapted to take the weight
  into account. At the time of this writing almost all classifiers in
  the popular Python package scikit-learn do include a class weight
  parameter; but gradient boosting machines are one of the important
  exceptions. A classifier-independent approach is to incorporate the
  cost into a boosting approach
  \citep{domingos1999metacost, sun2007cost}.
\item
  \textbf{Sampling approaches}. These strategies aim at creating a
  different training set for the classifier, where the imbalance of the
  two classes is significantly reduced. The sampling strategies can be
  further divided into two families.

  \begin{itemize}
  \tightlist
  \item
    \textbf{Undersampling} approaches reduce the number of majority
    class elements from the training set. The selection of the data to
    be removed from the training set can be at random
    \citep{drummond2003c4} or it can be informed one some measure of how
    '' important'' or ``non-important'' are the negative data points
    \citep{smith2014instance} or how ``noisy'' or ``non-noisy'' are the
    negative example data points \citep{kubat1997addressing}.
    \citet{imblearn} distinguish between three classes of undersampling:

    \begin{itemize}
    \tightlist
    \item
      \emph{prototype generator} where, for example, the negative class
      is clustered using k-means into \(n\) groups, and the centers of
      the clusters are the new negative data points, and the rest of the
      negative data is discarted;
    \item
      \emph{controlled} undesampling techniques, where the final
      quantity of negative examples can be controlled or set by the
      user;
    \item
      \emph{cleaning} undersampling techniques, where only negative data
      with some ``noisy'' or ``unimportant'' characteristic are
      eliminated from the majority side. An important concept in
      cleaning undersamplers is a \emph{Tomek link} - a pair of data
      from different classes that are each others closest neighbors. One
      simple cleaning undersampler is the Tomek undersampler
      (\texttt{tomekUS}) that removes all majority members of Tomek
      links. Notice that the algorithm is iterative, since removing the
      majority member of a Tomek link may reveal another Tomek link with
      the same minority member, but there is no way to control how many
      majority class data remains after the iterative removal.
    \end{itemize}
  \item
    \textbf{Oversampling} on the other hand, creates new
    \citep{chawla2002smote, he2008adasyn} or repeat positive examples,
    increasing their numbers in the training set. As an example of
    creating new positive examples, the Synthetic Minority Oversampling
    Technique or SMOTE \citep{chawla2002smote} selects a positive
    example, and from its \(k\) nearest positive neighbours will select
    a random one. Then it will create a new positive data point in a
    random position the line segment between these two positives
    examples.
  \end{itemize}
\item
  \textbf{Ensemble methods}. One can create different under-sampled
  training sets and train the base classifier on each of these different
  sets and combine the result of the different classifiers, for example
  by voting, into a single result. This is a bagging ensemble, based on
  different under-sampled training set. One can also create a bagging
  ensemble based on over-sampled training sets. Finally, one can also
  follow a boosting approach to ensembles, where the training set of
  each boost is a different under-sampled (or over-sampled) set.
\item
  \textbf{Specialised classifiers}. There has been some research on
  modifying known algorithms to better deal with imbalanced problems.
  For example, modifying SVM \citep{tang2008svms}, or random forests
  \citep{chen2004using}, or extreme learning machines
  \citep{zong2013weighted}, among others. Also, what in general could be
  considered an ensemble algorithm for any base classifier, is usually
  presented or used in the literature with a fixed base classifier,
  usually a decision tree. In this case we will also treat this as a
  specialized algorithm. For example EasyEnsemble
  \citep{liu2009exploratory} is a bagging of boosted base classifiers,
  but the literature only considers the case where the base classifier
  is a decision tree, and thus we will consider it as a specialised
  classifier and not as a general way of combining any base classifier.
  Similarly, RUSBoost \citep{seiffert2010rusboost} could be considered
  as the general construction procedure of using AdaBoost to construct a
  boosting ensemble and using random undersampling (thus the ``RUS'' in
  the name) on the training set of each boost. But in the literature it
  is always assumed that the base classifier is a decision tree, and
  thus we will consider it as a specialised classifier for imbalanced
  problems.
\end{itemize}

The families above are not the only way of classifying the strategies
into different families, nor it is the case that all strategies fall
into the classes above. For example, \citet{book} place the imbalanced
data strategies into four main classes: sampling and cost-based (as
above) but also kernel-based approaches \citep{wu2005kba, liu2007face},
which are more specific to SVM classifiers, and one-class approaches
\citep{raskutti2004extreme} which use one-class classifiers to learn
only the minority class. Kernel based approaches will be included into
our specialized algorithms class, but one-class approaches (for example,
\citep{raskutti2004extreme}) are not covered by our taxonomy.

\citet{estabrooks2004multiple} classify imbalance strategies broadly on
internal and external approaches. Internal approaches aim at changing
the formulation of a particular classifier so that it can deal better
with imbalanced data. Cost-sensitive approaches, such as class weight
and kernel-based approaches discussed in the previous paragraph are
internal. External approaches assume an unmodified base classifier, and
construct a larger classifier by usually combining different instances
of the base classifier trained on different samples of the original
data. All sampling approaches are external, as is the MetaCost
\citep{domingos1999metacost} framework for a cost-sensitive approach, as
well as ensemble approaches.

Also, not all strategies proposed in the literature fall within one of
our families, as we mentioned above regarding the one-class approaches.
\citet{batista2004study} proposed a mixed strategy that combines
oversampling followed by a cleaning undersampling, which also is not
covered by our taxonomy.

This research attempts to answer the question of what to do with an
imbalanced data set from the point of view of a practitioner. We assume
that a practitioner:

\begin{enumerate}
\def\labelenumi{\arabic{enumi}.}
\item
  will have time and computer resources to test some but not a very
  large number of imbalanced strategies, different base classifiers, and
  different hyperparameters for both the strategies and the base
  classifiers;
\item
  prefers using already implemented approaches;
\item
  may use different metrics;
\item
  does not have a deep understanding of his or her own data or an
  understanding of the literature regarding solutions that deal with
  ``similar'' data.
\end{enumerate}

We understand a practitioner as someone who is tasked to construct a
workable classifier of a particular data set in weeks, as opposed to a
researcher that has to construct the best solution for a classification
problem in months or years. The practitioner will prefer off-the-shelf
classifiers and already implemented imbalance strategies, and will
likely not spend the effort to understand his own data in depth beyond
the fact that it is imbalanced. The practitioner will not have the time
or the tools to understand the intrinsic characteristics of the data set
(if there are small disjuncts, overlapping, or low density of the
positive class) that may have an impact on the efficiency of different
imbalance mitigation strategies.

Let us discuss each of those characteristics of the practitioner, and
their impact on the methodological approach taken by this research.

Algorithms in classification are usually dependent on hyperparameters;
an algorithm that performs better the other algorithms in a dataset
using a particular set of hyperparameter values may perform much worse
if the hyperparameters were changed. Thus, almost all practitioners know
that one has to perform some hyperparameter search to achieve the best
performance of any algorithm. Furthermore, a practitioner also knows
that there are very few guarantees that one algorithm will outperform
all the others for all data sets. Thus, we assume that a practitioner
has both the time and the computational resources to test different
algorithms with different hyperparameters for their particular problem.

The same is true for most imbalanced strategies which also have
hyperparameters that have to be searched for. But most imbalance
strategies have an added degree of freedom: the base classifier itself.
Sampling strategies alter the training data that will be fed to a base
classifier, and ensemble strategies construct ensembles of a base
classifier. Furthermore, the base classifier may have hyperparameters
itself. We assume that the practitioner will not only test multiple
strategies and their hyperparameters, and different base algorithms and
their hyperparameters.

There is a reasonable literature that random forests, gradient boosting
machines, and SVM with radial kernels are, in general, the best
classifiers for the majority of problems. There is no proof for that,
but empirical research on a large set of data sets
\citep{fernandez2014we, arxiv} allows one to be reasonably sure that
those three classifiers are certainly the first three that should be
tested on any problem. We will assume that the practitioner will test
these base classifiers. This choice of base classifiers (and how we will
analyze their results - see Section \ref{sec:statistical-analysis}) is
an important difference of this research from previous published
comparisons (Section \ref{sec:related-literature}).

The second characteristic is that the practitioner will prefer already
implemented and well establish solutions. At the time of the writing of
this research, we believe that at least among Python and R
implementations of machine learning frameworks, the most well
established set of tools to deal with imbalanced are the one provided by
the \textbf{imblearn} Python package \citep{imblearn}. Imblearn is
compatible with the scikit-learn Python framework for machine learning,
and implements a large set of techniques (oversampling, undersampling,
some ensemble techniques, and a few specialized algorithms). As a
consequence to this research, we tested most of the imbalanced
mitigation algorithms implemented in imblearn. The version used in this
research is 0.12.0.dev0.

We also assume that the practitioner will not have decision power on the
metric used. Metrics are an important issue regarding imbalanced data
sets. The main problem with the standard metric, accuracy -- the
proportion of correct predictions -- is that it assigns the same cost
for the mis-classifications of both positive and negative examples. As
we discussed above, an important aspect of imbalanced problems is that
there is some understanding that misclassifying a positive case (a false
positive) should be costlier than misclassifying a negative case (a
false negative). Different metrics have been proposed to ``balance'' the
mistakes of the classifier on both the positive and negative classes
\citep{japkowicz2013assessment}.

But we believe that the practitioner will have little control over which
metric to use and in this paper we will not advocate one or other
metrics as the ``correct ones''. We will compare the different
strategies using different metrics, in particular the ones listed as the
most common ones used in the literature on both techniques and
applications of imbalanced problems, according to
\citet{haixiang2017learning}. The metrics are, in order of frequency of
use in the literature: AUC, accuracy, gmean, f1, recall, specificity,
precision, followed by balanced accuracy and MCC. These metrics are
discussed in Section~\ref{sec:metrics} and all but specificity are
tested in this research. (Specificity is the accuracy of the majority
case which we do not believe is an interesting aspect of imbalanced
problems.)

Finally, there is a current understanding that the source of
difficulties regarding imbalanced datasets is not due to the skewed
distribution of the classes but stem from the intrinsic characteristics
discussed above \citep{lopez2013insight}. If that is true, the best
strategy to deal with a particular data set may depend on its particular
intrinsic characteristics. We assume that a practitioner will not have
the tools or the time to explore these issues regarding their particular
data set.

\hypertarget{related-literature}{%
\subsection{\texorpdfstring{Related literature
\label{sec:related-literature}}{Related literature }}\label{related-literature}}

The standard in any research in imbalanced problems, and in machine
learning in general, is that researchers that propose new algorithms or
methods would compare those algorithms with others across different data
sets, using one or sometimes a few metrics. But those comparisons are
potentially partial. The choice of the competing algorithms, the choice
of data sets used in the comparisons, and the choice of metrics are made
by the researchers themselves. This is the standard practice within the
research community as a way to provide evidence that one algorithm is a
contribution to the research front, and we are not criticising this
practice. We will call a comparison as (potentially more) impartial if
the author is not the researcher that proposed one of the algorithms
that perform well in the comparison. We will discuss some of the
relevant impartial comparisons among imbalance mitigation strategies.

\citet{smote-comparison} compares 85 different oversampling algorithms
on 104 binary imbalanced data sets, with imbalance rates from 1.8 to
130, with the majority of the data sets having IR below 20. They used
gmean, AUC, F1 score, and precision at top 20\%, which measures the
accuracy on the positive class of the top 20\% of the examples with
higher internal score (for the classifier). As base classifier they use
SVM, decision trees, kNN, and multi-layer perceptrons. When aggregating
the results for all data set, for all base classifiers, and for all 4
metrics, the paper recommends polynom-fit-SMOTE \citep{gazzah2008new},
ProWSyn \citep{barua2013prowsyn} and SMOTE-IPF \citep{saez2015smote} as
the best oversampling algorithms.

\citet{galar2012review} focus on ensemble-based strategies to deal with
imbalanced data. They compare 7 different proposals for cost-based
boosting where the weight update rule of AdaBoost is changed to take
into consideration the cost of making a minority class error; 4
different algorithms in the family of boosting based ensembles, where
sampling strategies of adding or removing data for the training set of
each boost is used; 4 different bagging-based ensembles, which use
sampling and bagging; and 2 algorithms classified as hybrid. They only
use C4.5 as the base classifier and AUC as the quality metric. They
conclude that SMOTEBagging, RUSBoost, and UnderBagging are the best
strategies in terms of the AUC and RUSBoost seems to be the less
computationally complex solution.

\citet{lopez2013insight} compare 7 oversampling strategies, 3
cost-sensitive learning, and 5 ensemble-based strategies, using three
base classifiers (kNN, C4.5, and SVM) and AUC as the metric. They first
compare the strategies within each family and then compare the best in
each family to each other. Within the oversampling strategies, they
conclude that SMOTE and SMOTE+ENN perform better. For the cost-sensitive
approaches, class weight is the winning strategy; and among
ensemble-based strategies, RUSBoost, SMOTEBagging, and EasyEnsemble
perform better. When comparing the winners among themselves and a
no-strategies approach (which we call baseline in this paper), they
concluded that, in general, class weight, SMOTE, and SMOTE+ENN perform
better than the other strategies (for AUC).

This paper only deals with binary imbalanced problems. There has been
recent literature on multiclass imbalanced literature
\citep{fernandez2013analysing, liao2008classification, wang2012multiclass, abdi2016combat}.
For these problems, some research follows the path of decomposing the
multiclass problem into either OVO or OVA binary imbalanced problems
\citep{fernandez2013analysing, liao2008classification} while other
research tackles the full multiclass problem directly
\citep{wang2012multiclass, abdi2016combat}.

\hypertarget{methods}{%
\section{\texorpdfstring{Methods
\label{sec:methods}}{Methods }}\label{methods}}

This section describes the details of the experiment performed in this
research. Section~\ref{sec:strategies} describes the strategies tested;
Section~\ref{sec:metrics} discusses the metrics used to evaluate the
quality of the classifier and Section \ref{sec:data-classifers}
discusses the datasets used in the evaluation.
Section~\ref{sec:statistical-analysis} discusses the statistical
analysis performed in partiuclar the inovative way this research deals
with base classifiers; Section \ref{sec:class} discusses the different
base classifiers and hyperparameters explored in this research;
Section~\ref{sec:experimental-setup} describes the experimental setup in
terms of cross-validation used. Finally and
Section~\ref{sec:reproducibility} discusses the public availability of
both data and programs to reproduce this research.

\hypertarget{strategies}{%
\subsection{\texorpdfstring{Strategies
\label{sec:strategies}}{Strategies }}\label{strategies}}

The choice of strategies selected for the comparisons was based on one
of our assumptions regarding the goals of a practitioner: that she will
prefer solutions that are already implemented and established. In this
line, we used the python package Imbalanced Learn \citep{imblearn}, and
tested some of the strategies implemented therein.

\begin{itemize}
\item
  \texttt{base} Baseline is the null case - the use of a single base
  classifier with equal weighting for the classes.
\item
  \texttt{cw} The Class Weight strategy weights the positive examples
  with a higher value so that erring the positive examples is costlier
  than negative examples. This is an internal strategy and so it depends
  on changing the base classifier appropriately. Fortunately, standard
  implementations of random forest, and SVM already accepts a class
  weight parameter. Unfortunately, at the time of the writing of this
  report, there is no widely used implementation of gradient boosting
  classifiers that accept class weight. So this is a low-cost
  alternative for the practitioner. We fixed the class weight for the
  negative class as 1 and for the positive as the inverse of the
  imbalance rate, that is, for an imbalance rate of 100, the positive
  class has a weight of 100, and the negative weight 1.
\item
  Under the general undersampling approach, we tested the algorithms
  below. The algorithms names are an abbreviation of the general known
  name of the algorithm with the \texttt{US} suffix, to indicate that
  the algorithm belongs to the undersampling family. The algorithms for
  the other families will also use suffixes: \texttt{OS} for
  oversampling, \texttt{SA} for specialized algorithms, \texttt{EN} for
  ensemble algorithms.

  \begin{itemize}
  \item
    Controlled undersamplers:

    \begin{itemize}
    \item
      \texttt{randomUS} Randomly remove negative examples from the
      training set. This is the standard controlled undersampling
      method.
    \item
      \texttt{nmissUS} Near Miss \citep{nearmiss} (We will not discuss
      the details of each algorithm.)
    \end{itemize}
  \item
    Cleaning undersamplers

    \begin{itemize}
    \item
      \texttt{tomekUS} Tomek undersampler \citep{tomek1976x}.
    \item
      \texttt{editUS} Edited Nearest Neighbours
      \citep{wilson1972asymptotic}
    \item
      \texttt{allknnUS} All K-Nearest Neighbors \citep{tomek1976x}
    \item
      \texttt{condensedUS} Condensed Nearest Neighbour
      \citep{hart1968condensed}
    \item
      \texttt{onesidedUS} One Sided Selection \citep{onesided}
    \item
      \texttt{ncleaningUS} Neighbourhood Cleaning Rule
      \citep{laurikkala2001improving}
    \item
      \texttt{ihardnessUS} Instance Hardness Threshold
      \citep{smith2014instance}
    \end{itemize}
  \item
    Prototype undersampler:

    \begin{itemize}
    \tightlist
    \item
      \texttt{clusterUS} Cluster Centroids \citep{zhang2010cluster}
    \end{itemize}
  \end{itemize}
\item
  \textbf{oversampling} Under the general oversampling approach we
  tested the following algorithms:

  \begin{itemize}
  \item
    \texttt{randomOS} Random oversampling is the process of randomly
    selecting with replacement a positive data point, and adding it
    again to the training set.
  \item
    \texttt{smoteOS} SMOTE or Synthetic Minority Over-sampling Technique
    \citep{chawla2002smote}.
  \item
    \texttt{adasynOS} Adaptive Synthetic (ADASYN) sampling
    \citep{he2008adasyn}
  \item
    \texttt{bordersmoteOS} Borderline-SMOTE \citep{han2005borderline}
  \item
    \texttt{svmsmoteOS} SVM-SMOTE \citep{nguyen2011borderline}
  \end{itemize}
\item
  \textbf{bagging ensemble}. We used a two simple strategies based on
  bagging-based ensemble of the base classifier. The strategies are
  \texttt{underbagEN} and \texttt{overbagEN}; UnderbagEN is also called
  underbagging or balanced bagging classifiers, and it uses a standard
  bagging approach to construct the ensemble, where each bag is balanced
  using random undersampling. OverbagEN is a standard bagging approach
  where each bag is balanced using random oversampling.
\item
  \textbf{specialised classifiers}

  \begin{itemize}
  \item
    \texttt{easySA} EasyEnsemble \citep{liu2009exploratory}
  \item
    \texttt{rusbSA} RUSBoost \citep{seiffert2010rusboost}
  \item
    \texttt{balrfSA} Balanced Random Forests \citep{chen2004using}
  \end{itemize}
\end{itemize}

\hypertarget{metrics}{%
\subsection{\texorpdfstring{Metrics
\label{sec:metrics}}{Metrics }}\label{metrics}}

This paper will focus on eight metrics:

\begin{itemize}
\item
  \textbf{accuracy} abbreviated as \emph{acc}
\item
  \textbf{area under the ROC curve} abbreviated as \emph{auc}
\item
  \textbf{balanced accuracy} abbreviated as \emph{bac}
\item
  \textbf{F1-measure} abbreviated as \emph{f1}
\item
  \textbf{G-mean} \citep{kubat1998machine} abbreviated as \emph{gmean}
\item
  \textbf{Matthew's correlation
  coefficient}\citep{matthews1975comparison} abbreviated as \emph{mcc}
\item
  \textbf{precision} abbreviated as \emph{prec}
\item
  \textbf{recall} abbreviated as \emph{rec}
\end{itemize}

Let us denote TP as the number of true positives, that is, the correct
positive predictions of the classifier; TN as the number of true
negatives; FP is the number of false positives, that is data that the
classifier predicted as positive which were, in fact, negative; and FN
as the number of false negatives. We remind the reader that in this
paper and usually in imbalanced problems, the minority class is the
positive class and the majority class is the negative class. Then:

\[\begin{aligned}
 \mbox{acc} &= \frac{TP+TN}{TP+FP+TN+FN} \\
\mbox{precision} &= \frac{TP}{TP+FP} \\ \mbox{recall} &=
\frac{TP}{TP+FN} \\ \mbox{specificity} &= \frac{TN}{TN+FP} \\
\mbox{f1} &= \frac{2}{\frac{1}{\mbox{precision}} +
\frac{1}{\mbox{recall}}} = \frac{2TP}{2TP+FP+FN} \\ \mbox{bac} & =
\frac{\mbox{recall} + \mbox{specificity} }{2} = \frac{\frac{TP}{TP+FN}
+ \frac{TN}{TN+FP}}{2}\\ \mbox{gmean} &= \sqrt{\mbox{recall} \times
\mbox{specificity} } = \sqrt{\frac{TP}{TP+FN} \times \frac{TN}{TN+FP}
} \\ \mbox{mcc} &= \frac{TP \times TN - FP \times
FN}{\sqrt{(TP+FP)(TP+FN)(TN+FP)(TN+FN)}} \\\end{aligned}\]

\emph{Accuracy} is the proportion of correct predictions. Notice that
TP+FN is the total number of positive examples, and thus \emph{recall}
is the accuracy of the positive examples (the proportion of true
predictions for the positive examples). Similarly, \emph{specificity} is
the accuracy for the negative cases.

\emph{Precision} also known as positive predicted value, measures the
proportion of data that is indeed positive (TP) among the cases that the
classifier determined to be positive (TP+FP); or in other words, how
much can one trust the classifier if it outputs a positive
classification.

Classifiers usually compute a score value for each data. Let us assume
that higher scores are associated with the positive class. Besides
computing the score, the classifier also defines a threshold, above
which it classifies the example as positive. Changing this threshold,
one can change which examples are classified as positive or negative.
Lowering the threshold will change the classification of some examples
that before were negative to positive. Let assume that some of these
were indeed positive (before they were a false negative) but not all of
them. Lowering the threshold caused an increase in the true positives
and thus an increase in recall. But the newly false positives decrease
specificity - the accuracy of negative cases. Therefore, by changing the
threshold one can trade \emph{recall} for \emph{specificity}, increasing
one while decreasing the other. Similarly, one can trade \emph{recall}
for \emph{precision} by changing the threshold.

The metrics \emph{bac}, \emph{f1} and \emph{gmean} attempt to balance
two of these conflicting/tradeable metrics. \emph{bac} balances the
accuracy of the positive cases (recall) with the accuracy of the
negative cases (specificity) by taking the arithmetic mean of both
figures. \emph{f1} computes the harmonic mean between \emph{recall} and
\emph{precision}. \emph{gmean} computes the geometric mean between
\emph{recall} and \emph{specificity}.

MCC is the Pearson correlation between the predicted and the correct
classes. The usual Pearson correlation formula:

\[\label{eq:3}
  \rho(X,Y) = \frac{E[XY]-E[X]E[Y]}{\sqrt{E[X^2]-E[X]^2} \sqrt{E[Y^2]-E[Y]^2}}
\]

where \(X\) is the predicted class and \(Y\) is the correct class
reduces to the mcc formula above when one realizes that:

\begin{itemize}
\item
  \(X\) and \(Y\) are binary, and thus \(E[X^2] = E[X]\)
\item
  \(E[X]\) is the proportion of examples classified as positive
  (\((TP+FP)/N\)), where \(N = TP+FP+TN+FN\) is the total number of
  examples
\item
  \(E[Y]\) is the proportion of positive examples (\((TP+FN)/N\)),
\item
  \(E[XY]\) is the proportion of true positives (\(TP/N\))
\end{itemize}

Area under the curve (AUC) is defined in terms of the receiver operating
characteristic (ROC) curve, which is constructed by plotting the recall
and specificity (actually 1 - specificity) of the classifier for
different values of the score threshold discussed above. The ROC is the
convex hull of those points and the AUC is the area of that ROC.

A more intuitive definition is that the AUC is an estimate of the
probability that a random positive example will have a higher score than
a randomly selected negative example. Under this interpretation,
assuming that \(f(x)\) is the score for sample \(x\), \(X_{+}\) is the
set of positive examples, and \(X_{-}\) the set of negative samples,
then

\[\begin{aligned}
  \label{eq:2}
  auc &= \frac{1}{|X_{+}|\times |X_{-}| } \sum_{p \in X_{+}, n \in
        X_{-}} H( f(p) - f(n)) \end{aligned}
\]

where \(H(x) = +1\) if \(x>0\), and 0 otherwise, is the Heaviside step
function. The formula just counts the number of times the score of a
positive example is higher than the score of a negative example, divided
by the number of pairs.

There are other metrics proposed in the literature
\citep{scott2007performance, japkowicz2013assessment, branco2016survey}.
In this paper we focus on the eight metrics listed above: auc, acc, f1,
gmean, mcc, bac, prec, and rec.

We understand that accuracy is a metric that violates one of the
characteristics of an imbalance problem, the intuition that
misclassifying positive examples should be more costly that
misclassifying negative examples, and thus it is inappropriate as a
metric for an imbalanced problem. But given our goal answering the
practitioner's need for practical knowledge on how to solve an
imbalanced problem, we cannot disregard that they seem to use accuracy
with frequency.

\hypertarget{data}{%
\subsection{\texorpdfstring{Data
\label{sec:data-classifers}}{Data }}\label{data}}

This paper uses a standard set of data sets that has been used in many
imbalanced research. The data that we call \textbf{natural} are a set of
58 real life imbalanced data sets, with imbalance rate ranging from 2.5
to 160. The data is the first three blocks (``Imbalance ratio between
1.5 and 9'' and ``Imbalance ratio higher than 9'' Parts I and II) of the
Keel \citep{alcala2011keel} curated data sets on imbalance\footnote{The
  original three blocks from Keel contain 66 data sets. We removed from
  the analysis 8 data sets that would cause some of the algorithms not
  to run. They were ``shuttle-c2-vs-c4'',
  ``iris0'',``glass5'',``glass-0-4\_vs\_5'',
  ``glass-0-6\_vs\_5'',``ecoli-0-1-3-7\_vs\_2-6'',``glass-0-1-6\_vs\_5'',
  and ``new-thyroid2''.} .

For, the second data, which we call \textbf{manipulated}, we randomly
removed either majority or minority data in both the training and test
sets so that the resulting data sets would have specific IR of 3, 100,
or 300, if possible. The only restriction is that we limit the number of
minority data to be at least 8 for the training set and at least 2 for
the test set. All 58 datasets could be manipulated to an IR of 3, which
represent the low IR examples. Only two of the data sets could be
manipulated to an IR of 300, and other 7 could be manipulated for an IR
of 100. These 9 data sets represent the examples of high IR.

\hypertarget{statistical-analysis}{%
\subsection{\texorpdfstring{Statistical analysis
\label{sec:statistical-analysis}}{Statistical analysis }}\label{statistical-analysis}}

This paper aims to analyze various strategies within each family of
techniques to address imbalanced classification problems. However, the
implementation of nearly every strategy necessitates the configuration
of hyperparameters. To illustrate, for a controlled undersampling
strategy, a decision must be made concerning the final imbalance rate
post-undersampling. Besides this hyperparameter, there is a need to
select the base classifier (and its hyperparameters) that will process
the undersampled training data. Consequently, when comparing different
strategies, the choice of base classifier typically must be made, except
in the case of specialized algorithms, which are themselves the base
algorithm. A pertinent methodological issue is whether the base
algorithm is regarded as a \emph{nuisance factor}, in the broad sense,
or as a \emph{hyperparameter}.

Hyperparameters represent decisions made by the practitioner after
testing various alternatives. When executing a comparison of strategies,
treating the resulting imbalance ratio (IR) for a controlled
undersampler as a hyperparameter implies the assumption that the
practitioner, for their specific data set, will conduct a search for
possible values for this hyperparameter and select the optimal one. A
nuisance factor, on the other hand, is a decision significant for the
strategy's evaluation, but the researcher conducting the comparison does
not know which value the practitioner will opt for. Therefore, from the
viewpoint of the researcher making the comparison, the solution lies in
employing a sufficiently broad range of possible values for the nuisance
factor and averaging the results derived from all of them.

Prior research comparing various strategies has often treated the base
algorithm as a nuisance factor. This study, however, advocates for
treating the base algorithm as a hyperparameter, positing that
practitioners will not only explore different values for the strategy's
hyperparameters but will also experiment with different base algorithms
and their respective hyperparameters, and choose the best one.

In order to compare strategies, we employ a two-step procedure.
Initially, strategies within a single family are compared, with the
top-performing strategy designated as the representative for that
family. We then carry out a comparison of the representatives across
different families. The families of baseline, class weight, and ensemble
each comprise a single strategy, so these strategies naturally serve as
their respective representatives. However, for the families of
undersampling, oversampling, and specialized algorithms, a comparison of
all strategies within the family is performed to identify the
representative for each metric.

Regarding statistical analysis, we largely adhere to Demsar's procedure
\citep{demvsar2006statistical} for comparing multiple classifiers across
multiple data sets, albeit with modifications as suggested by
\citet{garcia2008extension} and \citet{garcia2010advanced}. Notably,
instead of using the Nemenyi test as advocated by Demsar, we employ a
pairwise Wilcoxon test on the ranks of each classifier, with a
multi-comparison adjustment of the p-value in accordance with Bergmann
and Hommel \citep{bergmann1988improvements}. For instances where more
than nine classifiers were compared, we utilized the Holm procedure
\citep{holm1979simple}. The computations were performed using the R
package scmamp \citep{calvo2016scmamp}.

\begin{figure}
\begin{floatrow}
\ffigbox{%
  \includegraphics[scale=0.5]{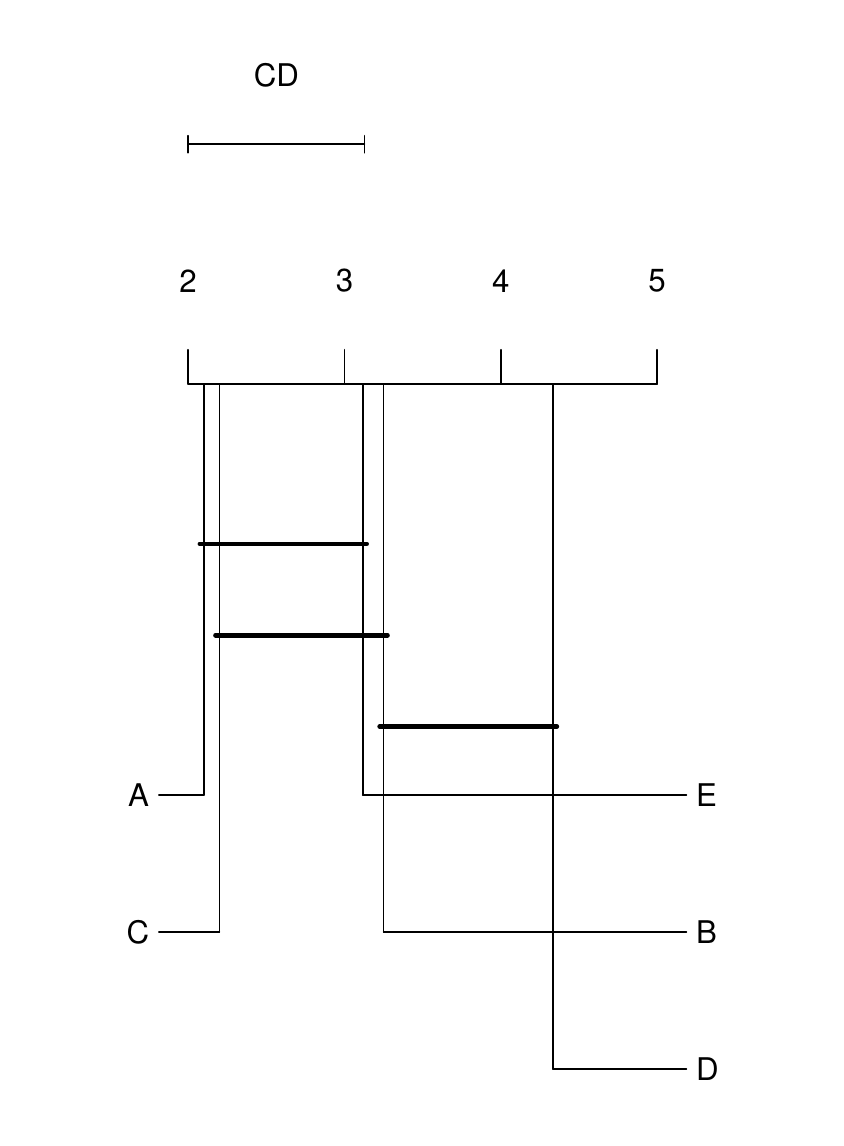}%
}{%
  \caption{The CD plot of the base results}\label{fig:demsar1}.%
}
\capbtabbox{%
\centering
\begin{tabular}{l>{\ttfamily}l}
\hline
algorithm & sig \\
\hline
  A   &  a \\
  C   &  ab \\
  E   &  ab \\
  B   &  ~bc \\
  D   &  ~~c \\
\hline
\end{tabular}
}{%
  \caption{The table representation of the CD plot}\label{tab:demsar1}%
}
\end{floatrow}
\end{figure}

Demsar suggested a graphical representation to visualize the resulting
comparisons, as depicted in Figure \ref{fig:demsar1}. In contrast, we
have chosen a more compact representation, utilizing a table to
concurrently display the ranking of different strategies and the
significance of the differences \citep{piepho2004algorithm}. Each letter
in Table \ref{tab:demsar1} corresponds to a bar in the plot, indicating
the lack of statistically significant differences among the algorithms.

Section \ref{sec:class} discusses the base classifiers, their
hyperparameters, and the hyperparameters for the strategies tested in
this research. Once we have a best combination of the base classifier,
the base classifier's hyperparameters, and the strategy's
hyperparameters, we will compare this strategy with the others. Section
\ref{sec:experimental-setup} discusses how the data is split into
training and test, so the quality of a particular strategy (using a
particular metric) is measured in the (multiple) test sets.

\hypertarget{classifiers-and-hyperparameters}{%
\subsection{\texorpdfstring{Classifiers and hyperparameters
\label{sec:class}}{Classifiers and hyperparameters }}\label{classifiers-and-hyperparameters}}

The base classifiers used are:

\begin{itemize}
\item
  random forest (rf). The hyperparameter range is: number of trees
  \(\in \{100, 500, 1000\}\).
\item
  SVM with RBF kernel (svm). The hyperparameters are:
  \(C \in \{2^{-5}, 2^{0}, 20^{5}, 20^{10},2^{15}\}\) and
  \(\gamma \in \{ 2^{-15},2^{-10},2^{-5},2^{0}, ,2^{5}\}\).
\item
  Gradient boosting machines (gbm). Hyperparameters: learning rate
  \(\in \{0.03, 0.1, 0.3\}\), max depth \(\in \{2, 3, 5\}\).
\end{itemize}

The specialised algorithms also have hyperparameters:

\begin{itemize}
\item
  \texttt{rusbSA}: the number of trees is a hyperparameter whose
  possible values are nboost \(\in \{ 10,20, 50, 100\}\).
\item
  \texttt{easySA}: The hyperparameter: number of estimators
  \(\in \{3, 7, 10, 20, 30\}\)
\item
  \texttt{balrfSA}: The hyperparameter: number of trees
  \(\in \{10, 50, 100, 500, 1000\}\)
\end{itemize}

The \texttt{underbagEN} and \texttt{overbagEN} in the ensemble family
also have a hyperparameter: number of bags \(\in \{5, 10, 20, 30\}\).

Finally, some of the over- and undersampling algorithms allow to control
for the final imbalance rate after the sampling. For these strategies
(\texttt{randomOS}, \texttt{smoteOS}, \texttt{bordersmoteOS},
\texttt{adasynOS}, \texttt{svmsmoteOS}, \texttt{nmissUS},
\texttt{ihardnessUS}, and \texttt{clusterUS}) the final imbalance rate
is also a hyperparameter, with the values in
\(\{ 0.2, 0.4, 0.6, 0.8, 1.0 \}\). We did not set the other
hyperparameters of the strategies and base classifiers.

\hypertarget{experimental-setup}{%
\subsection{\texorpdfstring{Experimental setup
\label{sec:experimental-setup}}{Experimental setup }}\label{experimental-setup}}

For each data set, we randomly selected 20\% of the data set as the test
set. The proportion of classes was the same in both the training and
test set but we did not impose any other constraints on the partitioning
(against the warnings of \citep{lopez2014importance}).

Let us call \emph{solution} a combination of a strategy and a base
classifier, or the specialised classifiers by themselves. We performed a
3-fold cross-validation on the training set to select the best set of
hyperparameters for the solution. For a particular metric, say auc, we
tested 10 random samples of the hyperparameters for each solution. The
combination of hyperparameters that maximized the mean auc over the
three folds was selected. Then the solution was trained on the whole
training set, and the auc on the test set was measured. The same process
was repeated for the other metrics.

We repeated the process (starting at the random selection of the test
set) 3 times, that is, we have three measures of the quality of the
solution, which we averaged to have a better estimate. Formally, we used
a 3-repetition of a 20\% holdout procedure with a nested 3-fold for
selecting the hyperparameters.

\hypertarget{sec:reproducibility}{%
\subsection{Reproducibility}\label{sec:reproducibility}}

The imbalanced data sets, the programs that run the experiments, the
results of all experiments, and the R program to perform the analysis
are available at \url{https://figshare.com/s/96b3d7f8d3f74de4b6e3}.

\hypertarget{sec:results}{%
\section{Results}\label{sec:results}}

\hypertarget{natural-data-sets}{%
\subsection{\texorpdfstring{Natural data sets
\label{sec:r1}}{Natural data sets }}\label{natural-data-sets}}

The statistical comparisons between strategies within the OS, US and SA
families are displayed in \ref{ap1}. The summary of those results are:

For \textbf{acc}:

\begin{itemize}
\item
  undersampling: \texttt{tomekUS} is the algorithm with best mean rank,
  but it is not significantly different than \texttt{onesidedUS},
  \texttt{ncleaningUS}, \texttt{clusterUS}, \texttt{editnnUS} and
  \texttt{allknnUS}.
\item
  oversampling: best algorithm \texttt{randomOS} but none of the other
  algorithms are significantly different from it.
\item
  specialised algorithms: \texttt{rusbSA} is the best algorithm, but not
  significantly better than \texttt{balrfSA}.
\item
  ensemble: \texttt{overbagEN} is significantly better than
  \texttt{underbagEN}.
\end{itemize}

For \textbf{auc}:

\begin{itemize}
\item
  undersampling; \texttt{editUS} is the algorithm with the best mean
  rank, but it is not significantly different than the other
  undersampling algorithm with the exception of \texttt{ihardnessUS}.
\item
  oversampling: best strategy is \texttt{svmsmoteOS}, but not
  significantly better than any of the other algorithm.
\item
  specialised algorithms: \texttt{balrfSA} is the best strategy and
  significantly better than the other two.
\item
  ensemble: \texttt{overbagEN} is not significantly better than
  \texttt{underbagEN}.
\end{itemize}

For \textbf{bac}:

\begin{itemize}
\item
  undersampling: \texttt{randomUS} is the best ranked algorithm, but not
  significantly better than \texttt{ihardnessUS}, \texttt{clusterUS},
  and \texttt{editnnUS}.
\item
  oversampling: best algorithm is \texttt{adasynOS}, but not
  significantly better than any of the others.
\item
  specialised algorithms: \texttt{balrfSA} is the best algorithm,
  significantly better than any of the others.
\item
  ensemble: \texttt{underbagEN} is significantly better than
  \texttt{overbagEN}.
\end{itemize}

For \textbf{f1}:

\begin{itemize}
\item
  undersampling: best strategy: \texttt{ncleaningUS} but not
  significantly better than the other strategies except
  \texttt{ihardnessUS}, \texttt{condensedUS}, and \texttt{nmissUS}.
\item
  oversampling: best strategy \texttt{svmsmoteOS}, but not significantly
  better than any of the other strategies.
\item
  specialised algorithms: \texttt{balrfSA} is the best strategy,
  significantly better than any of the other strategies.
\item
  ensemble: \texttt{overbagEN} is significantly better than
  \texttt{underbagEN}.
\end{itemize}

For \textbf{gmean}:

\begin{itemize}
\item
  undersampling: best algorithm \texttt{onesidedUS}, but not
  significantly better than \texttt{tomekUS}, \texttt{ncleaningUS},
  \texttt{editnnUS}, \texttt{allknnUS}, and \texttt{condensedUS}.
\item
  oversampling: best algorithm \texttt{svmsmoteOS}, but not
  significantly better any other strategy except \texttt{bordersmoteOS}.
\item
  specialised algorithms: \texttt{balrfSA} is the best algorithm,
  significantly better than the others.
\item
  ensemble: \texttt{overbagEN} is not significantly better than
  \texttt{underbagEN}.
\end{itemize}

For \textbf{mcc}:

\begin{itemize}
\item
  undersampling: best algorithm \texttt{ihardnessUS} but not
  significantly better than \texttt{randomUS}, \texttt{clusterUS},
  \texttt{allknnUS} and \texttt{editnnUS}.
\item
  oversampling: best algorithm \texttt{adasynOS}, but not significantly
  better than any of the others, except \texttt{randomOS}.
\item
  specialised algorithms: \texttt{balrfSA} is the best algorithm but not
  significantly better than \texttt{the\ others}easySA`.
\item
  ensemble: \texttt{underbagEN} is significantly better than
  \texttt{overbagEN}.
\end{itemize}

For \textbf{prec}:

\begin{itemize}
\item
  undersampling: best algorithm \texttt{tomekUS} but not significantly
  better than \texttt{ncleaningUS}, \texttt{onesidedUS},
  \texttt{clusterUS}, and \texttt{allknnUS}.
\item
  oversampling: best algorithm \texttt{randomOS}, but not significantly
  better than the other algorithms except \texttt{svmsmoteOS}.
\item
  specialised algorithms: \texttt{rusbSA} is the best algorithm but not
  significantly better than \texttt{balrfSA}.
\item
  ensemble: \texttt{overbagEN} is significantly better than
  \texttt{underbagEN}.
\end{itemize}

For \textbf{rec}:

\begin{itemize}
\item
  undersampling: best algorithm \texttt{clusterUS} but not significantly
  better than \texttt{ihardnessUS}, \texttt{randomUS},
  \texttt{condensedUS}, and \texttt{nmissUS}.
\item
  oversampling: best algorithm \texttt{adasynOS}, but not significantly
  better than the other algorithms.
\item
  specialised algorithms: \texttt{balrfSA} is the best algorithm,
  significantly better than the others.
\item
  ensemble: \texttt{underbagEN} is significantly better than
  \texttt{overbagEN}
\end{itemize}

\hypertarget{comparison-of-the-best-solutions}{%
\subsubsection{\texorpdfstring{Comparison of the best solutions
\label{sec:comp-best-solut}}{Comparison of the best solutions }}\label{comparison-of-the-best-solutions}}

\begin{figure}[ht]
  \centering
  \includegraphics{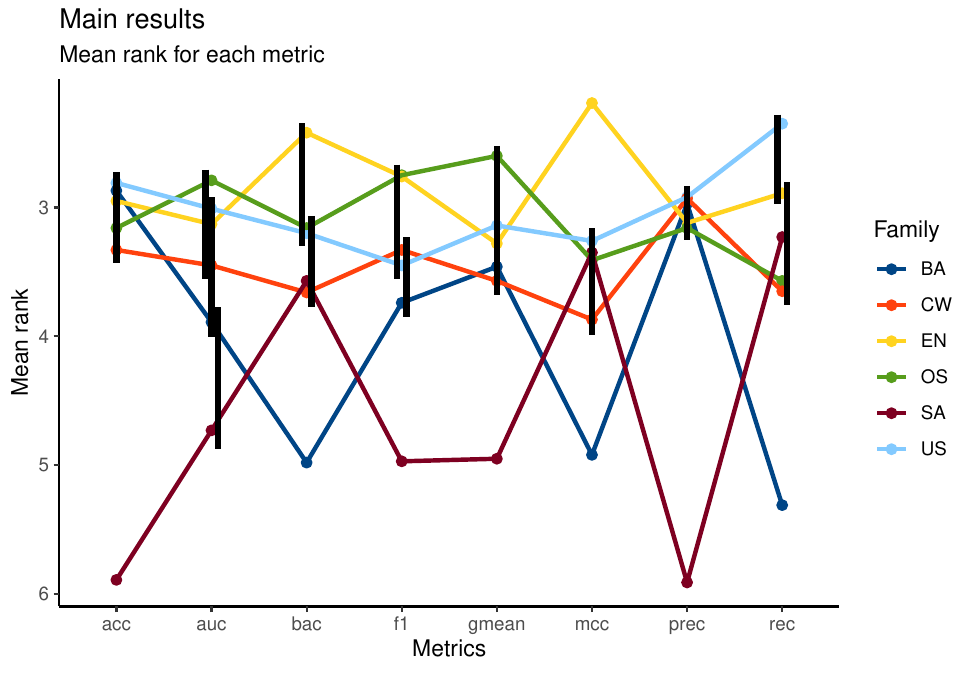}
  \caption{The mean rank of the best representative of each family. Mean rank is better the lower number; vertical dark line indicates that the differences are not statistically significant; horizontal axis indicates the metric.}
  \label{fig:main}
\end{figure}

We now compare the representative algorithm from each family of
oversampling, undersampling, ensemble, and specialized algorithms with
the baseline, and class weight, for each metric. The results are
displayed in Table \ref{tab:r1}. Figure \ref{fig:main} report visually
the results from Table \ref{tab:r1}

\begin{table}[ht]
\centering
\scalebox{0.9}{
\begin{tabular}{ll>{\ttfamily}l|ll>{\ttfamily}l|ll>{\ttfamily}l|}
  \hline
family & rank & sig & family & rank & sig & family & rank & sig \\ 
  \hline 
\rowcolor{lightgray} \multicolumn{ 3 }{c|}{ acc }&\multicolumn{ 3 }{c|}{ auc }&\multicolumn{ 3 }{c|}{ bac } \\
 \hline
  US & 2.81 & a~ & OS & 2.79 & a~~ & EN & 2.42 & a~~ \\ 
  BA & 2.87 & a~ & US & 3.01 & ab~ & OS & 3.16 & ab~ \\ 
  EN & 2.95 & a~ & EN & 3.13 & ab~ & US & 3.20 & ab~ \\ 
  OS & 3.16 & a~ & CW & 3.45 & ab~ & SA & 3.57 & ~b~ \\ 
  CW & 3.33 & a~ & BA & 3.89 & ~bc & CW & 3.66 & ~b~ \\ 
  SA & 5.89 & ~b & SA & 4.73 & ~~c & BA & 4.98 & ~~c \\ 
   \hline
\rowcolor{lightgray} \multicolumn{ 3 }{c|}{ f1 }&\multicolumn{ 3 }{c|}{ gmean }&\multicolumn{ 3 }{c|}{ mcc } \\
 \hline
  OS & 2.75 & a~~ & OS & 2.60 & a~ & EN & 2.19 & a~~ \\ 
  EN & 2.76 & a~~ & US & 3.14 & a~ & US & 3.26 & ~b~ \\ 
  CW & 3.33 & ab~ & EN & 3.28 & a~ & SA & 3.35 & ~b~ \\ 
  US & 3.45 & ab~ & BA & 3.46 & a~ & OS & 3.41 & ~b~ \\ 
  BA & 3.74 & ~b~ & CW & 3.57 & a~ & CW & 3.87 & ~b~ \\ 
  SA & 4.97 & ~~c & SA & 4.95 & ~b & BA & 4.92 & ~~c \\ 
   \hline
\rowcolor{lightgray} \multicolumn{ 3 }{c|}{ prec }&\multicolumn{ 3 }{c|}{ rec } \\
 \cline{1-6}
  US & 2.92 & a~ & US & 2.35 & a~~ \\ 
  CW & 2.93 & a~ & EN & 2.89 & ab~ \\ 
  BA & 2.97 & a~ & SA & 3.23 & ~b~ \\ 
  EN & 3.12 & a~ & OS & 3.57 & ~b~ \\ 
  OS & 3.16 & a~ & CW & 3.65 & ~b~ \\ 
  SA & 5.91 & ~b & BA & 5.31 & ~~c \\ 
   \cline{1-6}
\end{tabular}
}
\caption{Comparison of the best representatives from each family. \emph{rank} is the mean rank of the best representative for that family and \emph{sig} is the indication of statistical non-significance.  }\label{tab:r1}
\end{table}

In summary:

\begin{itemize}
\item
  For \textbf{acc} there is no statistically significant difference
  among undersampling, baseline, ensemble, oversampling, and cost
  weights strategies. Specialized algorithms are significantly worse
  than the best families.
\item
  For \textbf{auc}, there is no statistically significant difference
  among oversampling, undersampling, ensemble, and cost weight
  strategies. Specialized algorithms is significantly worse than the
  other families.
\item
  For \textbf{bac}, ensemble, oversampling and undersampling perform
  better.
\item
  For \textbf{f1} oversampling, ensemble, class weight and undesampling
  perform better.
\item
  For \textbf{gmean}, oversampling, undersampling, ensemble, baseline,
  and class weight perform better, and specialised algorithms performs
  significantly worse than the others.
\item
  For \textbf{mcc} ensemble performs significantly better then the
  alternatives.
\item
  For \textbf{prec}, undersampling, class weight, baseline, ensemble,
  and oversampling perform better. Specialised algorithms perform
  significantly worse than the others.
\item
  and finally for \textbf{rec}, undersampling, and ensemble perform
  better. Baseline performs significantly worse than the others.
\end{itemize}

\hypertarget{sec:best-combination}{%
\subsection{Different imbalance rates}\label{sec:best-combination}}

\begin{figure}[h]
  \centering
  \includegraphics{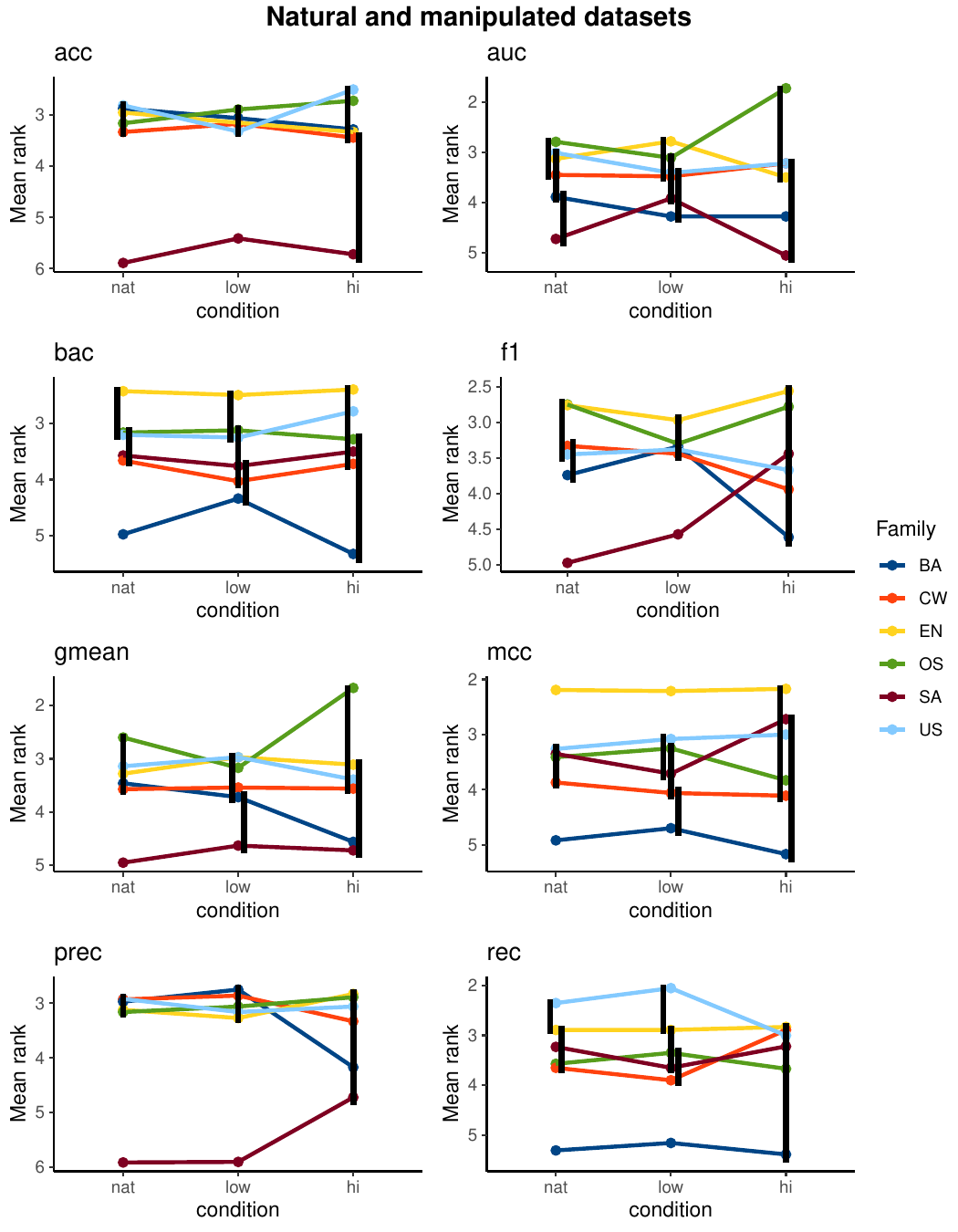}
  \caption{The families of the best strategies for the natural datasets, for the manipulated datasets with low IR (3 and 10) and the manipulated datasets with high IR (100 and 300).}
  \label{fig:lowhi}
\end{figure}

Figure \ref{fig:lowhi} displays the graphic result of the natural data
sets and the manipulated data sets for low (3) and high (100 and 300)
imbalance rates. For the high IR experiments, because of the smaller
number of data sets, the non-significance bars are longer. As expected
in most cases the families of strategies maintain their relative
positions. But there are some notable changes from low to high IR:

\begin{itemize}
\item
  for \emph{acc}, it is likely (but not demonstrated by our data) that
  OS and US will perform better in high IR than the other good
  strategies (BA, CW and EN).
\item
  for \emph{auc}, it is likely (but not demonstrated by our data) that
  OS should perform better for higher IR.
\item
  for \emph{gmean}, it is likely (but not demonstrated by our data) that
  OS should perform better for higher IR.
\end{itemize}

\hypertarget{discussion}{%
\section{\texorpdfstring{Discussion
\label{sec:discussion}}{Discussion }}\label{discussion}}

\hypertarget{dependency-on-the-quality-metric}{%
\subsection{Dependency on the quality
metric}\label{dependency-on-the-quality-metric}}

A critical observation from the results presented above is that the
choice of the optimal strategy is profoundly influenced by the metric
employed. To our knowledge, this interdependence has not been
highlighted in previous literature, and it bears significant
implications for both practitioners and researchers. Practitioners must
first determine the metric they will employ to assess the quality of the
classifier before experimenting with different imbalance strategies. For
instance, if the AUC metric is utilized, the baseline or class weight
strategies are likely to be the most effective. However, these
strategies perform poorly when the MCC metric is applied, with both
differences being statistically significant.

As discussed, this research will not provide practitioners with a
definitive guide on which metric is most advantageous or which is the
``correct'' one. We refer the reader to other resources that discuss the
various metrics: \citep{gu2009evaluation}, \citep{jeni2013facing}, and
\citep{japkowicz2013assessment}. Readers should also be mindful of works
such as \citep{hernandez2012unified} and \citep{hand2009measuring} that
establish a connection between performance metrics and anticipated
classification loss when operational conditions (like the ratio of
imbalance and the costs of errors for different classes) are unknown in
advance.

We believe that the strong dependence of the best strategy on the chosen
metric has not been a widely recognized phenomenon in the field. Very
few papers, and no comprehensive reviews of the area explicitly state
this. \citet{raeder2012learning} did briefly observe that certain
findings regarding the best classifier for imbalanced problems depend on
the evaluation metric, but this observation was used to advocate for the
superiority of one metric over others. \citet{zhu2017empirical} arrived
at a conclusion akin to our own but confined their analysis to churn
prediction problems. In their study, 24 strategies were tested on 11
churn-specific datasets, leading to the conclusion that ``The evaluation
metric has a high impact on the (measured, ranked) performance of
different techniques.''

The significance of the evaluation metric on the results should inspire
future researchers in this field to compare their novel algorithms and
strategies using a variety of metrics. A strategy that surpasses
alternatives using one metric might also excel with other metrics. By
providing comparisons across different metrics, the researcher can
deliver valuable insights to practitioners, aiding them in deciding
whether to test a specific algorithm or strategy once they have
determined the quality metric.

\hypertarget{base-classifier-as-hyperparameter-and-relation-to-previous-results}{%
\subsection{\texorpdfstring{Base classifier as hyperparameter and
relation to previous results
\label{basehyper}}{Base classifier as hyperparameter and relation to previous results }}\label{base-classifier-as-hyperparameter-and-relation-to-previous-results}}

The second important conclusion is that this research seems to
contradict the published results, in particular, the results that use
AUC as metric \citep{prati2015class, galar2012review, lopez2013insight}.
All of these papers show specialised algorithms (for example RUSBoost)
as a winning strategy, while for us, it is the worse family of
strategies.

\begin{table}

\caption{\label{tab:dt2} \label{dt2}  Table of comparisons for AUC, when using the hyperparameter method versus when using the nuisance method when dealing with base classifiers.}
\centering
\begin{tabular}[t]{lrllrl}
\toprule
\multicolumn{3}{c}{hyperparameter} & \multicolumn{3}{c}{nuisance} \\
\cmidrule(l{3pt}r{3pt}){1-3} \cmidrule(l{3pt}r{3pt}){4-6}
\textbf{algorithm} & \textbf{mean.rank} & \textbf{sig} & \textbf{algorithm} & \textbf{mean.rank} & \textbf{sig}\\
\midrule
\cellcolor{gray!6}{OS} & \cellcolor{gray!6}{2.79} & \cellcolor{gray!6}{a} & \cellcolor{gray!6}{OS} & \cellcolor{gray!6}{2.81} & \cellcolor{gray!6}{a}\\
US & 3.01 & ab & SA & 3.22 & a\\
\cellcolor{gray!6}{EN} & \cellcolor{gray!6}{3.13} & \cellcolor{gray!6}{ab} & \cellcolor{gray!6}{EN} & \cellcolor{gray!6}{3.26} & \cellcolor{gray!6}{a}\\
CW & 3.45 & ab & US & 3.59 & ab\\
\cellcolor{gray!6}{BA} & \cellcolor{gray!6}{3.89} & \cellcolor{gray!6}{bc} & \cellcolor{gray!6}{CW} & \cellcolor{gray!6}{3.66} & \cellcolor{gray!6}{ab}\\
\addlinespace
SA & 4.73 & c & BA & 4.47 & b\\
\bottomrule
\end{tabular}
\end{table}

A significant distinction between our study and previous research lies
in the treatment of the base algorithm as a hyperparameter in our
analysis (as detailed in Section \ref{sec:statistical-analysis}), while
other studies regard it as a nuisance factor. Besides this analytical
difference, past literature also often employs weak classifiers, or base
classifiers that are likely suboptimal for the data. Naturally, in a
nuisance factor analysis, where the average over possible values of the
base classifier is calculated, it would be reasonable to include simpler
algorithms in the average. Studies by \citet{prati2015class} and
\citet{lopez2013insight} include SVMs alongside less powerful
classifiers such as decision tree, rule induction, and naive Bayes, all
with fixed hyperparameters. In contrast, our research not only treats
the base algorithm as a hyperparameter (and seeks better hyperparameters
for the base classifiers themselves) but also uses ``stronger'' base
classifiers, or those more likely to perform well on the data.

To illustrate the impact of our approach, Table \ref{dt2} presents a
comparison of the statistical tests of AUC for our methodology, and the
results from including weaker base classifiers into the alternatives (in
this case, 1-NN and CART decision trees) and averaging the results for
all base classifiers, that is treating base classifiers as a nuisance
factor.

The revised results partially align with the conclusions of prior
research. For instance, \citet{galar2012review}, who employed only
decision trees as base classifiers, identified RUSBoost (SA) and
Underbagging (EN) as winning strategies, both of which rank as the top
two strategies in Table \ref{dt2} without any statistically significant
differences. These results, however, show less congruence with
\citet{lopez2013insight}, who found SMOTE (OS), RUSBoost (SA), and class
weight (CW) as winning strategies, obtained using k-NN, C4.5, and SVM
(all with fixed hyperparameters) as base classifiers.

Specialized algorithms, since they do not have base classifiers, benefit
a lot from the nuisance methodology. Strategies on the other families
will be weighted down by averaging their results using less powerful
base classifiers while specialized algorithms will not.

\hypertarget{testing-newer-algorithms}{%
\subsection{\texorpdfstring{Testing newer algorithms
\label{newalg}}{Testing newer algorithms }}\label{testing-newer-algorithms}}

A potential critique of this research is its reliance on the particular
strategies and algorithms tested, which may not represent the cutting
edge within each family. This section demonstrates that our main
findings remain relatively stable when we introduce newer algorithms.
Initially, we augmented our analysis by including only newer algorithms
from each family, followed by an inclusive approach where we
incorporated all new algorithms across families.

Within the OS family, we based our selection of newer (or more modern)
algorithms on an analysis by \citet{smote-comparison}, which compared 85
oversampling algorithms across three metrics: AUC, F1, and gmean. We
chose the top three algorithms for each metric. The new OS algorithms
tested included: polynom-fit-SMOTE \citep{gazzah2008new}, ProWSyn
\citep{barua2013prowsyn}, SMOTE-IPF \citep{saez2015smote}, Lee
\citep{lee2015over}, G-SMOTE \citep{sandhan2014handling}, LVQ-SMOTE
\citep{nakamura2013lvq}, Assembled-SMOTE \citep{zhou2013quasi}, and
SMOBD \citep{wang2012applying}. We extend our thanks to the authors for
their contributions to the implementation of these and other
oversampling algorithms.

For the ensemble family, our analysis so far considered two basic
algorithms: underbagging and overbagging. To examine new ensemble
strategies, we leveraged the implementation of various ensemble
alternatives by \citet{liu2021imbens}. Among these, we tested the
overboost ensemble, which is a boosting approach that incorporates
oversampling in its boosting stages, and the self-paced ensemble
\citep{selfpaced}, a novel proposal for addressing imbalanced datasets.

In the specialized algorithm category, we included the AdaCC
\citep{adacc} algorithm in our tests.

\begin{figure}[h]
 \centering
 \includegraphics{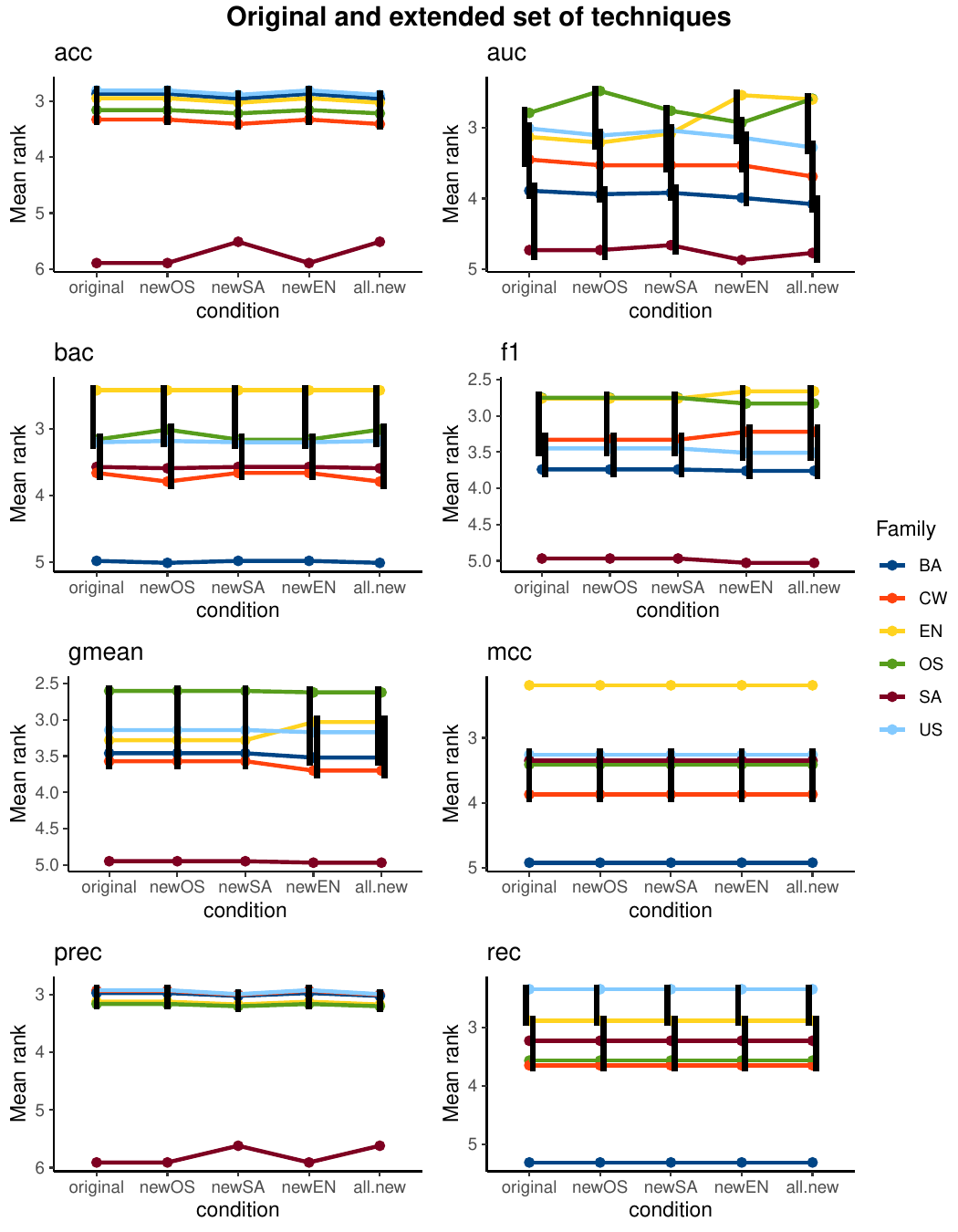}
 \caption{The comparison of our original results and results by adding newer algorithms in the OS, EN and SA families. newOS are the result of adding only newer OS strategies, and similarly for newEN and newSA.  .}\label{fig:news}
 \label{fig:newalgs}
\end{figure}

Our experimental process involved the sequential addition of new
strategies from one family at a time and then the simultaneous addition
of all the new strategies. Figure \ref{fig:news} compares the original
ranking of the families with the updated rankings after incorporating
the new strategies, culminating in the final entry where all new
strategies are included.

The integration of more sophisticated (or at least more contemporary)
oversampling (OS) algorithms does not alter the relative ranking of the
best-performing algorithms for each metric, with the exception of the
area under the curve (AUC) and balanced accuracy (BAC). For AUC, OS is
already the leading family of strategies (though not significantly
superior to some others); however, the introduction of newer algorithms
appears to further distinguish OS from the rest (yet without a
significant difference from ensemble (EN) and undersampling (US)).
Consequently, in the case of AUC, newer OS algorithms are likely to
result in performance improvements. Likewise, newer EN algorithms also
enhance that family's average ranking. For BAC, the latest OS algorithms
have shifted OS from the third to the second-best strategy family
(although not significantly different from the other two).

The introduction of more sophisticated EN algorithms seems to have a
greater impact on the EN family's relative ranking, especially for AUC
but also for the F1-score and geometric mean (gmean). Conversely, the
specialized algorithms (SA) are consistently outperformed by the other
families, and the addition of new SA algorithms does little to change
this.

Therefore, for all metrics, with the potential exception of AUC, one can
reasonably project the findings of this study to newer or forthcoming
algorithms---if this paper concludes that for Matthews correlation
coefficient (MCC), ensemble strategies are likely the best option based
on the tested strategies, it is safe to assume that a newly published OS
algorithm would not overturn that conclusion. More definitively, a
practitioner using MCC as their metric should focus on testing some of
the EN strategies discussed in this paper, and probably avoid expending
resources on implementing or testing a newly published OS algorithm.

AUC, however, appears to be a metric that is more susceptible to
fluctuations with the introduction of newer algorithms. In our
experiments, newer algorithms in the OS family did improve its relative
ranking, a trend that was also observed with the introduction of newer
EN algorithms. This indicates that some of these newer algorithms did
outperform their older counterparts. Therefore, a practitioner using AUC
as a metric should stay informed about newer algorithms, especially
among the families that generally perform well for this metric: OS, EN,
and US (for which we did not test newer algorithms).

\hypertarget{conclusions}{%
\section{\texorpdfstring{Conclusions
\label{sec:conclusions}}{Conclusions }}\label{conclusions}}

\hypertarget{how-a-practitioner-should-use-these-results}{%
\subsection{How a practitioner should use these
results}\label{how-a-practitioner-should-use-these-results}}

Once the practitioner selects a quality metric, they should refer to
Figure \ref{fig:main} or Table \ref{tab:r1} to identify the which
strategies are most effective according to that metric. Additionally,
practitioners must consider the data's imbalance rate (IR), especially
in cases of high IR. For these instances, the practitioner should review
the trends for high IR depicted in Figure \ref{fig:high_ir}. It is
important to note that these trends may not be definitive due to the
limited size of the datasets, which prevents achieving statistical
significance. Nonetheless, for those using metrics such as AUC or
G-mean, oversampling is highly recommended.

Upon determining the suitable strategy families, practitioners should
refer to section \ref{sec:r1} or Appendix A, which details the
top-performing algorithms within these families. Practitioners are
encouraged to test all, or at the very least, the majority of the
top-ranked algorithms from the chosen families.

Finally, practitioners should not overlook the importance of selecting
different base classifiers for strategies that require them and should
invest effort into hyperparameter tuning for both the algorithm and the
base classifiers. Section \ref{sec:class} provides a list of
hyperparameters and the values that have been utilized in this study.

\hypertarget{how-a-researcher-should-use-these-results}{%
\subsection{How a researcher should use these
results}\label{how-a-researcher-should-use-these-results}}

This paper offers two key recommendations for researchers who are
developing new algorithms to address imbalanced problems. The first, and
in the authors' view, the most pressing research direction, concerns the
alignment of strategies with various metrics. A pertinent question
arises: why do ensemble algorithms excel with the MCC metric, while
specialized algorithms perform poorly with the precision metric? A
deeper exploration into how these strategies correlate with specific
metrics could illuminate potential enhancements within those algorithmic
families. For instance, a better grasp of why the MCC metric aligns so
well with ensemble strategies might lead to advancements in ensemble
algorithms.

Secondly, researchers developing a new algorithm within a given family
should benchmark it against other algorithms in the same family,
particularly for the metrics where that family excels. If practitioners
are to adopt the new algorithm, it stands a better chance if it is among
the top performers for the metric of interest, rather than simply being
the best within its family.

\appendix

\hypertarget{results-on-the-natural-data-sets-each-family}{%
\section{\texorpdfstring{Results on the natural data sets: each family
\label{ap1}}{Results on the natural data sets: each family }}\label{results-on-the-natural-data-sets-each-family}}

\begin{table}[ht]
\centering
\scalebox{0.9}{
\begin{tabular}{ll>{\ttfamily}l|ll>{\ttfamily}l|ll>{\ttfamily}l|}
  \hline
algorithm & rank & sig & algorithm & rank & sig & algorithm & rank & sig \\ 
  \hline 
\rowcolor{lightgray} \multicolumn{ 3 }{c|}{ acc }&\multicolumn{ 3 }{c|}{ auc }&\multicolumn{ 3 }{c|}{ bac } \\
 \hline
  tomekUS & 4.04 & a~~ & editnnUS & 4.53 & a~ & randomUS & 3.53 & a~~~ \\ 
  onesidedUS & 4.33 & a~~ & allknnUS & 4.84 & a~ & ihardnessUS & 4.03 & ab~~ \\ 
  ncleaningUS & 4.66 & ab~ & ncleaningUS & 4.94 & a~ & clusterUS & 4.72 & abc~ \\ 
  clusterUS & 4.95 & ab~ & onesidedUS & 5.01 & a~ & editnnUS & 5.19 & abcd \\ 
  editnnUS & 5.11 & ab~ & tomekUS & 5.09 & a~ & allknnUS & 5.47 & ~bcd \\ 
  allknnUS & 5.50 & ab~ & randomUS & 5.65 & ab & ncleaningUS & 5.66 & ~bcd \\ 
  nmissUS & 6.13 & ~bc & clusterUS & 5.90 & ab & condensedUS & 6.23 & ~~cd \\ 
  randomUS & 6.20 & ~bc & condensedUS & 5.99 & ab & nmissUS & 6.48 & ~~cd \\ 
  condensedUS & 6.25 & ~bc & nmissUS & 6.14 & ab & tomekUS & 6.72 & ~~~d \\ 
  ihardnessUS & 7.84 & ~~c & ihardnessUS & 6.93 & ~b & onesidedUS & 6.96 & ~~~d \\ 
   \hline
\rowcolor{lightgray} \multicolumn{ 3 }{c|}{ f1 }&\multicolumn{ 3 }{c|}{ gmean }&\multicolumn{ 3 }{c|}{ mcc } \\
 \hline
  ncleaningUS & 4.35 & a~~ & onesidedUS & 4.11 & a~~ & ihardnessUS & 3.79 & a~~~~ \\ 
  tomekUS & 4.35 & a~~ & tomekUS & 4.29 & a~~ & randomUS & 4.05 & ab~~~ \\ 
  onesidedUS & 4.52 & a~~ & ncleaningUS & 4.36 & a~~ & clusterUS & 4.73 & abc~~ \\ 
  editnnUS & 4.84 & a~~ & editnnUS & 4.42 & a~~ & allknnUS & 5.20 & abcd~ \\ 
  allknnUS & 5.22 & ab~ & allknnUS & 4.47 & a~~ & editnnUS & 5.31 & abcd~ \\ 
  clusterUS & 5.34 & ab~ & condensedUS & 5.72 & ab~ & ncleaningUS & 5.70 & ~bcde \\ 
  randomUS & 5.66 & abc & randomUS & 6.22 & ~bc & nmissUS & 6.10 & ~~cde \\ 
  ihardnessUS & 6.63 & ~bc & clusterUS & 6.23 & ~bc & condensedUS & 6.15 & ~~cde \\ 
  condensedUS & 6.67 & ~bc & ihardnessUS & 7.33 & ~bc & tomekUS & 6.78 & ~~~de \\ 
  nmissUS & 7.42 & ~~c & nmissUS & 7.84 & ~~c & onesidedUS & 7.19 & ~~~~e \\ 
   \hline
\rowcolor{lightgray} \multicolumn{ 3 }{c|}{ prec }&\multicolumn{ 3 }{c|}{ rec } \\
 \cline{1-6}
  tomekUS & 3.58 & a~~~~ & clusterUS & 3.48 & a~~~ \\ 
  ncleaningUS & 3.94 & ab~~~ & ihardnessUS & 3.52 & a~~~ \\ 
  onesidedUS & 4.23 & ab~~~ & randomUS & 3.68 & a~~~ \\ 
  clusterUS & 5.04 & abc~~ & condensedUS & 4.27 & ab~~ \\ 
  allknnUS & 5.32 & abc~~ & nmissUS & 4.45 & ab~~ \\ 
  editnnUS & 5.44 & ~bc~~ & allknnUS & 5.72 & ~bc~ \\ 
  randomUS & 5.68 & ~bcd~ & editnnUS & 6.20 & ~~c~ \\ 
  nmissUS & 6.49 & ~~cde & ncleaningUS & 7.24 & ~~cd \\ 
  condensedUS & 7.33 & ~~~de & onesidedUS & 8.18 & ~~~d \\ 
  ihardnessUS & 7.95 & ~~~~e & tomekUS & 8.26 & ~~~d \\ 
   \cline{1-6}
\end{tabular}
}
\caption{Undersampling algorithms: results of the statistical test.} 
\end{table}

\begin{table}[ht]
\centering
\scalebox{0.9}{
\begin{tabular}{ll>{\ttfamily}l|ll>{\ttfamily}l|ll>{\ttfamily}l|}
  \hline
algorithm & rank & sig & algorithm & rank & sig & algorithm & rank & sig \\ 
  \hline 
\rowcolor{lightgray} \multicolumn{ 3 }{c|}{ acc }&\multicolumn{ 3 }{c|}{ auc }&\multicolumn{ 3 }{c|}{ bac } \\
 \hline
  randomOS & 2.84 & a & svmsmoteOS & 2.84 & a & adasynOS & 2.53 & a \\ 
  smoteOS & 2.95 & a & smoteOS & 2.85 & a & smoteOS & 2.96 & a \\ 
  svmsmoteOS & 2.99 & a & adasynOS & 2.95 & a & svmsmoteOS & 3.03 & a \\ 
  adasynOS & 3.04 & a & randomOS & 3.15 & a & bordersmoteOS & 3.14 & a \\ 
  bordersmoteOS & 3.17 & a & bordersmoteOS & 3.22 & a & randomOS & 3.34 & a \\ 
   \hline
\rowcolor{lightgray} \multicolumn{ 3 }{c|}{ f1 }&\multicolumn{ 3 }{c|}{ gmean }&\multicolumn{ 3 }{c|}{ mcc } \\
 \hline
  svmsmoteOS & 2.74 & a & svmsmoteOS & 2.55 & a~ & adasynOS & 2.44 & a~ \\ 
  randomOS & 2.91 & a & smoteOS & 2.70 & a~ & smoteOS & 2.99 & ab \\ 
  adasynOS & 3.07 & a & adasynOS & 2.98 & ab & bordersmoteOS & 3.07 & ab \\ 
  bordersmoteOS & 3.13 & a & randomOS & 3.16 & ab & svmsmoteOS & 3.09 & ab \\ 
  smoteOS & 3.15 & a & bordersmoteOS & 3.61 & ~b & randomOS & 3.41 & ~b \\ 
   \hline
\rowcolor{lightgray} \multicolumn{ 3 }{c|}{ prec }&\multicolumn{ 3 }{c|}{ rec } \\
 \cline{1-6}
  randomOS & 2.25 & a~ & adasynOS & 2.56 & a \\ 
  svmsmoteOS & 2.97 & ab & smoteOS & 2.97 & a \\ 
  smoteOS & 3.00 & ~b & bordersmoteOS & 2.98 & a \\ 
  bordersmoteOS & 3.23 & ~b & randomOS & 3.13 & a \\ 
  adasynOS & 3.55 & ~b & svmsmoteOS & 3.36 & a \\ 
   \cline{1-6}
\end{tabular}
}
\caption{Oversampling algorithms: results of the statistical test.} 
\end{table}

\begin{table}[ht]
\centering
\scalebox{0.9}{
\begin{tabular}{ll>{\ttfamily}l|ll>{\ttfamily}l|ll>{\ttfamily}l|}
  \hline
algorithm & rank & sig & algorithm & rank & sig & algorithm & rank & sig \\ 
  \hline 
\rowcolor{lightgray} \multicolumn{ 3 }{c|}{ acc }&\multicolumn{ 3 }{c|}{ auc }&\multicolumn{ 3 }{c|}{ bac } \\
 \hline
  rusbSA & 1.78 & a~ & balrfSA & 1.50 & a~ & balrfSA & 1.46 & a~~ \\ 
  balrfSA & 1.90 & a~ & easySA & 2.20 & ~b & easySA & 1.96 & ~b~ \\ 
  easySA & 2.33 & ~b & rusbSA & 2.30 & ~b & rusbSA & 2.59 & ~~c \\ 
   \hline
\rowcolor{lightgray} \multicolumn{ 3 }{c|}{ f1 }&\multicolumn{ 3 }{c|}{ gmean }&\multicolumn{ 3 }{c|}{ mcc } \\
 \hline
balrfSA & 1.43 & a~~ & balrfSA & 1.52 & a~ & balrfSA & 1.59 & a~ \\ 
  easySA & 2.09 & ~b~ & easySA & 2.16 & ~b & easySA & 1.94 & a~ \\ 
  rusbSA & 2.47 & ~~c & rusbSA & 2.32 & ~b & rusbSA & 2.47 & ~b \\ 
   \hline
\rowcolor{lightgray} \multicolumn{ 3 }{c|}{ prec }&\multicolumn{ 3 }{c|}{ rec } \\
 \cline{1-6}
  balrfSA & 1.76 & a~ & balrfSA & 1.49 & a~~ \\ 
  rusbSA & 1.89 & a~ & easySA & 1.89 & ~b~ \\ 
  easySA & 2.35 & ~b & rusbSA & 2.62 & ~~c \\ 
   \cline{1-6}
\end{tabular}
}
\caption{Specialized algorithms: results of the statistical test.} 
\end{table}

\begin{table}[ht]
\centering
\scalebox{0.9}{
\begin{tabular}{ll>{\ttfamily}l|ll>{\ttfamily}l|ll>{\ttfamily}l|}
  \hline
algorithm & rank & sig & algorithm & rank & sig & algorithm & rank & sig \\ 
  \hline 
\rowcolor{lightgray} \multicolumn{ 3 }{c|}{ acc }&\multicolumn{ 3 }{c|}{ auc }&\multicolumn{ 3 }{c|}{ bac } \\
 \hline
  overbagEN & 1.17 & a~ & overbagEN & 1.46 & a~ & underbagEN & 1.29 & a~ \\ 
  underbagEN & 1.83 & ~b & underbagEN & 1.54 & a~ & overbagEN & 1.71 & ~b \\ 
   \hline
\rowcolor{lightgray} \multicolumn{ 3 }{c|}{ f1 }&\multicolumn{ 3 }{c|}{ gmean }&\multicolumn{ 3 }{c|}{ mcc } \\
 \hline
  overbagEN & 1.28 & a~ & overbagEN & 1.41 & a~ & underbagEN & 1.24 & a~ \\ 
  underbagEN & 1.72 & ~b & underbagEN & 1.59 & a~ & overbagEN & 1.76 & ~b \\ 
   \hline
\rowcolor{lightgray} \multicolumn{ 3 }{c|}{ prec }&\multicolumn{ 3 }{c|}{ rec } \\
 \cline{1-6}
  overbagEN & 1.12 & a~ & underbagEN & 1.41 & a~ \\ 
  underbagEN & 1.88 & ~b & overbagEN & 1.59 & ~b \\ 
  \cline{1-6}
\end{tabular}
}
\caption{Ensemble strategies: results of the statistical test} 
\end{table}

\clearpage

\hypertarget{results-on-new-algorithms}{%
\section{\texorpdfstring{Results on new algorithms
\label{apnew}}{Results on new algorithms }}\label{results-on-new-algorithms}}

\begin{table}[ht]
\centering
\scalebox{0.65}{
  \begin{tabular}{ll>{\ttfamily}l|ll>{\ttfamily}l|ll>{\ttfamily}l|ll>{\ttfamily}l|ll>{\ttfamily}l|}
    \multicolumn{ 3 }{c}{original} &  \multicolumn{ 3 }{c}{newOS} &  \multicolumn{ 3 }{c}{newSA} &  \multicolumn{ 3 }{c}{newEN} &  \multicolumn{ 3 }{c}{all new} \\
  \hline
algorithm & rank & sig & algorithm & rank & sig & algorithm & rank & sig & algorithm & rank & sig & algorithm & rank & sig \\ 
  \hline 
\rowcolor{lightgray} \multicolumn{ 15 }{c|}{ acc } \\
 \hline
  US & 2.81 & a~ & US & 2.81 & a~ & US & 2.89 & a~ & US & 2.81 & a~ & US & 2.89 & a~ \\ 
  BA & 2.87 & a~ & BA & 2.87 & a~ & BA & 2.96 & a~ & BA & 2.87 & a~ & BA & 2.96 & a~ \\ 
  EN & 2.95 & a~ & EN & 2.95 & a~ & EN & 3.03 & a~ & EN & 2.95 & a~ & EN & 3.03 & a~ \\ 
  OS & 3.16 & a~ & OS & 3.16 & a~ & OS & 3.22 & a~ & OS & 3.16 & a~ & OS & 3.22 & a~ \\ 
  CW & 3.33 & a~ & CW & 3.33 & a~ & CW & 3.41 & a~ & CW & 3.33 & a~ & CW & 3.41 & a~ \\ 
  SA & 5.89 & ~b & SA & 5.89 & ~b & SA & 5.51 & ~b & SA & 5.89 & ~b & SA & 5.51 & ~b \\ 
   \hline
\rowcolor{lightgray} \multicolumn{ 15 }{c|}{ auc } \\
 \hline
  OS & 2.79 & a~~ & OS & 2.48 & a~~ & OS & 2.76 & a~~ & EN & 2.54 & a~~~ & OS & 2.59 & a~~ \\ 
  US & 3.01 & ab~ & US & 3.11 & ab~ & US & 3.04 & ab~ & OS & 2.93 & ab~~ & EN & 2.60 & a~~ \\ 
  EN & 3.13 & ab~ & EN & 3.21 & ab~ & EN & 3.08 & ab~ & US & 3.14 & abc~ & US & 3.28 & ab~ \\ 
  CW & 3.45 & ab~ & CW & 3.53 & ~b~ & CW & 3.53 & ab~ & CW & 3.53 & ~bc~ & CW & 3.69 & ~b~ \\ 
  BA & 3.89 & ~bc & BA & 3.94 & ~bc & BA & 3.92 & ~bc & BA & 3.99 & ~~c~ & BA & 4.08 & ~bc \\ 
  SA & 4.73 & ~~c & SA & 4.73 & ~~c & SA & 4.66 & ~~c & SA & 4.87 & ~~~d & SA & 4.77 & ~~c \\ 
   \hline
\rowcolor{lightgray} \multicolumn{ 15 }{c|}{ bac } \\
 \hline
  EN & 2.42 & a~~ & EN & 2.42 & a~~ & EN & 2.42 & a~~ & EN & 2.42 & a~~ & EN & 2.42 & a~~ \\ 
  OS & 3.16 & ab~ & OS & 3.01 & ab~ & OS & 3.16 & ab~ & OS & 3.16 & ab~ & OS & 3.01 & ab~ \\ 
  US & 3.20 & ab~ & US & 3.18 & ab~ & US & 3.20 & ab~ & US & 3.20 & ab~ & US & 3.18 & ab~ \\ 
  SA & 3.57 & ~b~ & SA & 3.59 & ~b~ & SA & 3.57 & ~b~ & SA & 3.57 & ~b~ & SA & 3.59 & ~b~ \\ 
  CW & 3.66 & ~b~ & CW & 3.79 & ~b~ & CW & 3.66 & ~b~ & CW & 3.66 & ~b~ & CW & 3.79 & ~b~ \\ 
  BA & 4.98 & ~~c & BA & 5.01 & ~~c & BA & 4.98 & ~~c & BA & 4.98 & ~~c & BA & 5.01 & ~~c \\ 
   \hline
\rowcolor{lightgray} \multicolumn{ 15 }{c|}{ f1 } \\
 \hline
  OS & 2.75 & a~~ & OS & 2.75 & a~~ & OS & 2.75 & a~~ & EN & 2.66 & a~~ & EN & 2.66 & a~~ \\ 
  EN & 2.76 & a~~ & EN & 2.76 & a~~ & EN & 2.76 & a~~ & OS & 2.83 & a~~ & OS & 2.83 & a~~ \\ 
  CW & 3.33 & ab~ & CW & 3.33 & ab~ & CW & 3.33 & ab~ & CW & 3.22 & ab~ & CW & 3.22 & ab~ \\ 
  US & 3.45 & ab~ & US & 3.45 & ab~ & US & 3.45 & ab~ & US & 3.51 & ab~ & US & 3.51 & ab~ \\ 
  BA & 3.74 & ~b~ & BA & 3.74 & ~b~ & BA & 3.74 & ~b~ & BA & 3.76 & ~b~ & BA & 3.76 & ~b~ \\ 
  SA & 4.97 & ~~c & SA & 4.97 & ~~c & SA & 4.97 & ~~c & SA & 5.03 & ~~c & SA & 5.03 & ~~c \\ 
   \hline
\rowcolor{lightgray} \multicolumn{ 15 }{c|}{ gmean } \\
 \hline
  OS & 2.60 & a~ & OS & 2.60 & a~ & OS & 2.60 & a~ & OS & 2.62 & a~~ & OS & 2.62 & a~~ \\ 
  US & 3.14 & a~ & US & 3.14 & a~ & US & 3.14 & a~ & EN & 3.03 & ab~ & EN & 3.03 & ab~ \\ 
  EN & 3.28 & a~ & EN & 3.28 & a~ & EN & 3.28 & a~ & US & 3.17 & ab~ & US & 3.17 & ab~ \\ 
  BA & 3.46 & a~ & BA & 3.46 & a~ & BA & 3.46 & a~ & BA & 3.52 & ab~ & BA & 3.52 & ab~ \\ 
  CW & 3.57 & a~ & CW & 3.57 & a~ & CW & 3.57 & a~ & CW & 3.70 & ~b~ & CW & 3.70 & ~b~ \\ 
  SA & 4.95 & ~b & SA & 4.95 & ~b & SA & 4.95 & ~b & SA & 4.97 & ~~c & SA & 4.97 & ~~c \\ 
   \hline
\rowcolor{lightgray} \multicolumn{ 15 }{c|}{ mcc } \\
 \hline
  EN & 2.19 & a~~ & EN & 2.19 & a~~ & EN & 2.19 & a~~ & EN & 2.19 & a~~ & EN & 2.19 & a~~ \\ 
  US & 3.26 & ~b~ & US & 3.26 & ~b~ & US & 3.26 & ~b~ & US & 3.26 & ~b~ & US & 3.26 & ~b~ \\ 
  SA & 3.35 & ~b~ & SA & 3.35 & ~b~ & SA & 3.35 & ~b~ & SA & 3.35 & ~b~ & SA & 3.35 & ~b~ \\ 
  OS & 3.41 & ~b~ & OS & 3.41 & ~b~ & OS & 3.41 & ~b~ & OS & 3.41 & ~b~ & OS & 3.41 & ~b~ \\ 
  CW & 3.87 & ~b~ & CW & 3.87 & ~b~ & CW & 3.87 & ~b~ & CW & 3.87 & ~b~ & CW & 3.87 & ~b~ \\ 
  BA & 4.92 & ~~c & BA & 4.92 & ~~c & BA & 4.92 & ~~c & BA & 4.92 & ~~c & BA & 4.92 & ~~c \\ 
   \hline
\rowcolor{lightgray} \multicolumn{ 15 }{c|}{ prec } \\
 \hline
  US & 2.92 & a~ & US & 2.92 & a~ & US & 2.99 & a~ & US & 2.92 & a~ & US & 2.99 & a~ \\ 
  CW & 2.93 & a~ & CW & 2.93 & a~ & CW & 3.00 & a~ & CW & 2.93 & a~ & CW & 3.00 & a~ \\ 
  BA & 2.97 & a~ & BA & 2.97 & a~ & BA & 3.02 & a~ & BA & 2.97 & a~ & BA & 3.02 & a~ \\ 
  EN & 3.12 & a~ & EN & 3.12 & a~ & EN & 3.17 & a~ & EN & 3.12 & a~ & EN & 3.17 & a~ \\ 
  OS & 3.16 & a~ & OS & 3.16 & a~ & OS & 3.20 & a~ & OS & 3.16 & a~ & OS & 3.20 & a~ \\ 
  SA & 5.91 & ~b & SA & 5.91 & ~b & SA & 5.62 & ~b & SA & 5.91 & ~b & SA & 5.62 & ~b \\ 
   \hline
\rowcolor{lightgray} \multicolumn{ 15 }{c|}{ rec } \\
 \hline
  US & 2.35 & a~~ & US & 2.35 & a~~ & US & 2.35 & a~~ & US & 2.35 & a~~ & US & 2.35 & a~~ \\ 
  EN & 2.89 & ab~ & EN & 2.89 & ab~ & EN & 2.89 & ab~ & EN & 2.89 & ab~ & EN & 2.89 & ab~ \\ 
  SA & 3.23 & ~b~ & SA & 3.23 & ~b~ & SA & 3.23 & ~b~ & SA & 3.23 & ~b~ & SA & 3.23 & ~b~ \\ 
  OS & 3.57 & ~b~ & OS & 3.57 & ~b~ & OS & 3.57 & ~b~ & OS & 3.57 & ~b~ & OS & 3.57 & ~b~ \\ 
  CW & 3.65 & ~b~ & CW & 3.65 & ~b~ & CW & 3.65 & ~b~ & CW & 3.65 & ~b~ & CW & 3.65 & ~b~ \\ 
  BA & 5.31 & ~~c & BA & 5.31 & ~~c & BA & 5.31 & ~~c & BA & 5.31 & ~~c & BA & 5.31 & ~~c \\ 
   \hline
\end{tabular}
}
\caption{Introduction of more modern strategies: results of the statistical tests.} 
\end{table}

\end{document}